\documentclass{article}
\usepackage{microtype}
\usepackage{graphicx}
\usepackage{booktabs}
\usepackage{hyperref}

\usepackage[preprint]{icml2026}
\usepackage{amsmath}
\usepackage{amssymb}
\usepackage{mathtools}
\usepackage[capitalize,noabbrev]{cleveref}
\newcommand{\mc}{\textsc{MoldCast}}
\newcommand{\sd}[1]{{\scriptstyle #1}}

\hypersetup{pdfsubject={}}

\icmltitlerunning{\textsc{MoldCast}: Contradiction Retrieval with a Single Inner Product}

\begin{document}

\twocolumn[
  \icmltitle{\textsc{MoldCast}: Contradiction Retrieval with a Single Inner Product}

  \begin{icmlauthorlist}
    \icmlauthor{Debayan Banerjee}{leu}
    \icmlauthor{Ricardo Usbeck}{leu}
  \end{icmlauthorlist}
  \icmlaffiliation{leu}{Leuphana University, L{\"u}neburg, Germany}
  \icmlcorrespondingauthor{Debayan Banerjee}{debayan.banerjee@leuphana.de}

  \icmlkeywords{contradiction retrieval, dense retrieval, inner product search, benchmarks}
  \vskip 0.3in
]

\printAffiliationsAndNotice{}

\begin{abstract}
Contradiction retrieval is a task that asks a system to find, in a large corpus, the passages that contradict an input query statement. Past work has argued that cosine similarity cannot express this relation. We present \mc, a contradiction detector whose score is a single inner product, so that any standard vector index can apply it to a whole corpus. \mc\ describes each passage by learned slots, designed so that two passages clash when their values differ in slots to which both commit. On the standard benchmarks, \mc\ performs on par with SparseCL, the current method, with a 40\% smaller index. However, we find that these benchmarks can largely be solved from the passage alone. We therefore build a new benchmark from real Wikipedia revisions that removes this shortcut. On it, \mc\ ranks contradictions above their confounders three times as often as SparseCL. An analysis shows that its values act as fingerprints of the numbers a sentence mentions. Code and data: \url{https://github.com/semantic-systems/moldcast}.

\end{abstract}

\section{Introduction}
\label{sec:intro}

Many systems need to check new information against what they already store. A fact-checking database receives a new claim. A knowledge base ingests a sentence extracted from text. In each case, the system needs to find the stored passages that \emph{contradict} the new statement. \citet{xu2025contradiction} formulated this problem as contradiction retrieval. Given a query, the task is to rank a large corpus so that the passages contradicting the query come first. Consider a query and two stored passages:

\begin{quote}
\small
\textbf{Query:} The town had 1,200 residents in 2010.\\
\textbf{Paraphrase:} In 2010, the town's population was 1,200.\\
\textbf{Contradiction:} The town had 1,400 residents in 2010.
\end{quote}

Only the third passage should be retrieved. We call the passage to retrieve the \emph{target}. A passage that is very similar to the query but agrees with it, like the paraphrase, is a \emph{confounder}: by similarity alone, it looks as good a match as the target. Yet it differs from the query in a single number, and it is at least as similar to the query as the paraphrase is. Dense retrieval therefore struggles with this task. Cosine similarity cannot express it even in principle, because similarity is transitive. The query and its paraphrase both contradict the third passage, but they do not contradict each other. Still, all three lie close together in the embedding space.

\begin{figure*}[t]
\centering
\includegraphics[width=\textwidth]{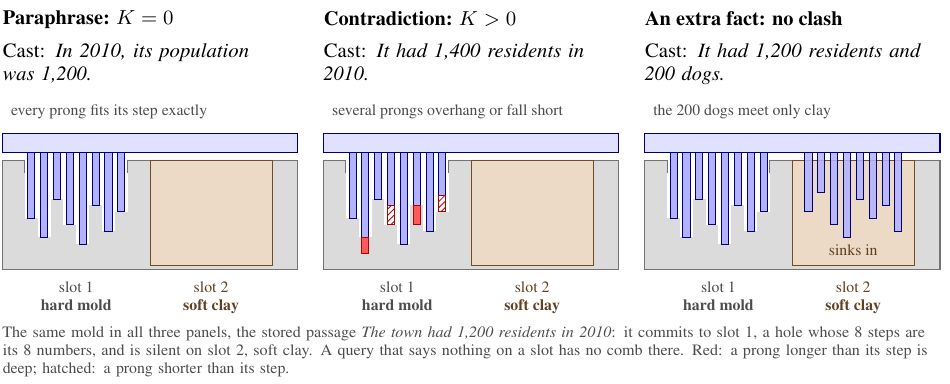}
\caption{\mc's clash for two slots of eight numbers, the $2\times 8$ configuration (illustrative values). All three panels show the same mold, the stored passage: it commits to slot~1, a hole whose bottom has eight steps, one for each number of the slot's value, and is silent on slot~2, which is soft clay. A query is a cast with a comb on each slot it commits to, whose eight prongs are as long as its numbers. Left: a paraphrase's prongs fit every step ($K=0$). Middle: a contradiction shifts several numbers of slot~1's value; each prong that is too long (red) or too short (hatched) adds its squared mismatch, so the slot's clash is the squared distance $\lVert v^q_c - v^d_c\rVert^2$. Right: a query adding a fact the stored passage is silent about (200 dogs) commits to slot~2 too, and its comb sinks into the clay without clashing: the extra fact neither supports nor contradicts the stored passage. In the model, commitments lie between 0 and 1, like the firmness of the material; trained models commit on every slot (\cref{sec:analysis}). What each slot captures is learned.}
\label{fig:mold}
\end{figure*}

We present \mc, a contradiction detector whose score \emph{is} an inner product. \mc\ describes each passage by a few learned slots. Each slot holds a commitment, which says whether the passage fills this slot, and a value, which says what it fills it with. We can picture a stored passage as a mold, with a hole wherever it commits to a slot. A query is then a cast, whose pegs either fit those holes or clash with them (\cref{fig:mold}). Two passages clash when both commit to the same slot with different values. The squared difference of their values can be expanded exactly into an asymmetric inner product. One side of this product is a vector stored once per passage, and the other is a vector built from the query. As a result, a passage and its paraphrase can be close while both clash with a contradiction, so the transitivity argument does not apply. In addition, any inner-product index can score the whole corpus exactly, without a second reranking stage.

We evaluate \mc\ in two rounds. In the first round, we compare it with SparseCL \citep{xu2025contradiction}, the current contradiction-retrieval method, on SparseCL's own benchmarks. There, the two methods perform on par (\cref{sec:round1}). While analysing these results, we found something we had not set out to measure. Much of what both methods gain over cosine is available from the passage alone, without comparing it with the query. This is because the generated contradictions in these benchmarks can be recognised by their style. These benchmarks therefore cannot show how well a method compares a query with a passage. In the second round, we build a new benchmark from real Wikipedia revisions in the VitaminC dataset \citep{schuster2021vitaminc}. In this benchmark, every passage is the contradiction of one query and the confounder of another (\cref{fig:symmetric}). No preference over passages alone can therefore solve it. We then compare the two methods again (\cref{sec:round2}).

Our contributions are the following:
\begin{enumerate}
\item \mc, a novel contradiction-retrieval model whose score is an inner product. Any vector index can therefore apply it to a whole corpus. Its index is 40\% smaller than SparseCL's (\cref{sec:method}).
\item The finding that, on the standard benchmarks, much of a trained method's gain survives without the query. This holds for \mc\ and SparseCL alike, and we show it with a query-free control (\cref{sec:round1}).
\item A benchmark built from VitaminC, on which a score that looks only at passages cannot separate a contradiction from its confounder.
\end{enumerate}

\section{Motivation: Contradiction Checking in Human Comprehension}
\label{sec:motivation}

People face a version of this problem whenever they read or hear something. A new statement is, in effect, checked against much of what they already know. Conflicts with world knowledge register in the brain within about 400\,ms \citep{hagoort2004integration}. This validation appears to be routine rather than deliberate \citep{richter2009you}. Models of reading describe it in two steps. Related memories are first activated by resonance, and the new information is then checked against them \citep{myers1998accessing,obrien2016coherence}. This is fast retrieval of potential conflicts from a large store, followed by a check. We use no brain or behavioural data, and we make no claim about how the brain does this. \mc\ is offered only as a computational model that is fast and scales to large stores (Appendix~\ref{app:human}).

\section{\mc}
\label{sec:method}

\begin{figure}[t]
\centering
\includegraphics[width=\columnwidth]{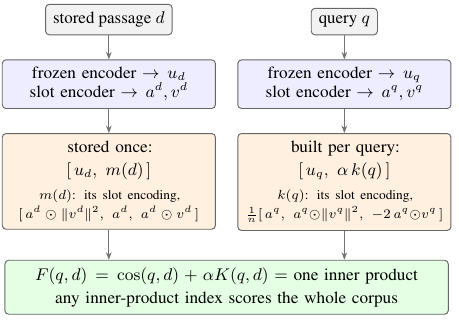}
\caption{\mc\ for contradiction retrieval. A passage is encoded once into its cosine embedding $u_d$ and its slot encoding $m(d)$: its commitments and values from the slot encoder, arranged so that the clash is an inner product (\cref{eq:inner}). A query becomes one vector in the same way, and the score of every stored passage, cosine plus the weighted clash, is one inner product.}
\label{fig:pipeline}
\end{figure}

\paragraph{Slots.} \mc\ keeps a frozen cosine embedding of each passage. Next to it, a trainable encoder describes each passage by $n$ slots. Slot $c$ holds a commitment $a_c \in (0,1)$ and a value $v_c \in \mathbb{R}^k$ (a $k$-valued vector). The commitment says whether the passage fills the slot, and the value says what it fills it with. What each slot captures is learned during training, not assigned in advance. In the picture of \cref{fig:mold}, a stored passage is a mold. It has one hole for each slot it commits to, and the bottom of the hole has one step for each number of the slot's value. Where the passage is silent, the mold is soft clay. A query is a cast, with one comb per slot whose prongs are as long as its numbers.

\paragraph{Clash.} We write $q$ for the query and $d$ for a stored passage. A superscript says whose slots we mean: $a^q_c$ and $v^q_c$ are the commitment and the value of the query in slot $c$, and $a^d_c$ and $v^d_c$ are those of the passage. Two passages clash when both commit to a slot but hold different values there. We measure this by the clash
\begin{equation}
K(q,d) = \frac{1}{n}\sum_{c=1}^{n} a^q_c\, a^d_c\, \bigl\lVert v^q_c - v^d_c \bigr\rVert^2 .
\label{eq:clash}
\end{equation}
Each term of the sum concerns one slot $c$. The product $a^q_c a^d_c$ is close to 1 only when both texts commit to the slot, so a slot that either text leaves silent contributes almost nothing. The squared distance $\lVert v^q_c - v^d_c\rVert^2$ measures how much the two values differ. It is 0 when they are equal, and it grows with their difference. Dividing by $n$ averages the terms over the slots.

The retrieval score adds the clash, weighted by $\alpha$, to the cosine similarity of the two frozen embeddings:
\begin{equation}
F(q,d) = \cos(q,d) + \alpha\, K(q,d).
\label{eq:score}
\end{equation}
The weight $\alpha$ is chosen on the development split. The two terms have different roles. Cosine finds the passages that are about the same thing as the query, and the clash decides which of them disagree with it.

Why can this score express contradiction when cosine cannot? With cosine, a high score means that two passages are close, and closeness carries over: if $a$ is close to $b$ and $b$ is close to $c$, then $a$ is close to $c$. In the example of the introduction, the contradiction must score high with the query, and also with the paraphrase, which contradicts it too. All three would then be close, and the paraphrase would score as high as the contradiction. The clash works the other way round. A high clash means that two passages hold values that are far apart, and being far apart does not carry over. The query and its paraphrase can both be far from the contradiction while holding the same value, so that the clash between them is 0.

\paragraph{One inner product.} We now show that the clash of \cref{eq:clash} can be written as an inner product. For each slot, we first expand the squared distance, where $v^q_c \cdot v^d_c$ is the dot product of the two values:
\[
\bigl\lVert v^q_c - v^d_c\bigr\rVert^2 = \lVert v^q_c\rVert^2 + \lVert v^d_c\rVert^2 - 2\, v^q_c \cdot v^d_c .
\]
Substituting this into the clash splits the term of each slot into three parts:
\begin{align*}
K(q,d) = \frac{1}{n}\sum_{c=1}^{n} \Bigl[\, & a^q_c \bigl(a^d_c \lVert v^d_c\rVert^2\bigr) \\
{}+{} & \bigl(a^q_c \lVert v^q_c\rVert^2\bigr)\, a^d_c \\
{}-{} & 2\bigl(a^q_c v^q_c\bigr)\cdot\bigl(a^d_c v^d_c\bigr) \Bigr].
\end{align*}
In each part, the first factor depends only on the query, and the second factor only on the stored passage. We collect the query factors of all slots into one vector $k(q)$, and the passage factors into one vector $m(d)$, in the same order. The clash is then exactly their inner product:
\begin{equation}
\begin{aligned}
K(q,d) &= \bigl\langle k(q),\, m(d)\bigr\rangle,\\
m(d) &= \bigl[\, a^d \odot \lVert v^d\rVert^2,\; a^d,\; a^d \odot v^d \,\bigr],\\
k(q) &= \tfrac{1}{n}\bigl[\, a^q,\; a^q \odot \lVert v^q\rVert^2,\; -2\, a^q \odot v^q \,\bigr].
\end{aligned}
\label{eq:inner}
\end{equation}
Here $a^d$ is the vector of the passage's $n$ commitments, and $v^d$ is the list of its $n$ values. $\lVert v^d\rVert^2$ is the vector of the squared lengths of these values, with one number per slot. The symbol $\odot$ multiplies two vectors element by element, slot by slot, and the square brackets join the three parts into one long vector. The stored vector $m(d)$ therefore has $n + n + nk = n(k+2)$ numbers, and the query vector $k(q)$ has the same size. If we append the cosine embedding to both vectors, the full score $F$ also becomes one inner product (\cref{fig:pipeline}). Any inner-product index can therefore score the whole corpus exactly, without a shortlist. With 16 slots of 8 numbers, a passage stores $768 + 16\times 10 = 928$ numbers. With 2 slots of one number, it stores 774. Note that the stored map $m$ and the query map $k$ are different. This is what makes $K(x,x)=0$ possible, so that a passage never clashes with itself.

\paragraph{Training.} Each training example is a triple. It contains a query, its contradiction, and a similar passage that agrees with the query. The contradiction must outscore the other contradictions and similar passages in the batch under the score $w_K K + w_{\cos}\cos$. This is a contrastive objective. Both weights are learned, and $w_{\cos}$ starts at 50, which corresponds to a temperature of 0.02. Known false negatives are masked out. The cosine encoder is frozen, and cosine prefers the similar passage. The only way to rank the contradiction first is therefore a large clash with the contradiction and a small clash with the passage that agrees. The slot encoder is fine-tuned from the same encoder as the frozen cosine model. Further details are in Appendix~\ref{app:details}.

\section{The Baseline: SparseCL}
\label{sec:baseline}

SparseCL \citep{xu2025contradiction} introduced contradiction retrieval as a task, and it is the method we compare with. SparseCL also keeps the frozen cosine embedding. It adds a second encoder $s$, which is trained so that a passage and its contradiction differ in only a few dimensions. A passage is scored by
\begin{equation}
F(q,d) = \cos(q,d) + \alpha\, H\bigl(s(d) - s(q)\bigr),
\label{eq:sparsecl}
\end{equation}
where $H$ is the Hoyer sparsity \citep{hoyer2004nonnegative} of the difference between the two sparse embeddings. For a vector $x$ with $D$ dimensions,
\begin{equation}
H(x) = \frac{\sqrt{D} - \lVert x\rVert_1 / \lVert x\rVert_2}{\sqrt{D} - 1},
\label{eq:hoyer}
\end{equation}
where $\lVert x\rVert_1$ is the sum of the absolute values of its entries and $\lVert x\rVert_2$ is its Euclidean length. The sparse embeddings have $D = 768$ dimensions. $H$ is 1 when the difference is concentrated in a single dimension, and 0 when it is spread evenly over all dimensions. The idea is that a paraphrase rewords a passage, which changes many dimensions a little, while a contradiction changes one fact, which should change only a few dimensions, but strongly. A high $H$ thus means that two passages are the same except for one specific thing. SparseCL trains the encoder $s$ so that this holds: differences to contradictions become sparse, and differences to paraphrases dense. The sparse encoder is trained with a contrastive objective of the same form as \mc's, with the Hoyer sparsity in place of the clash and a temperature of 0.02. The weight $\alpha$ is tuned on the development split. SparseCL's argument against cosine is the transitivity argument of the introduction. Its sparsity score escapes this argument, because a passage and its paraphrase are expected to differ densely, while a passage and its contradiction differ sparsely.

Because $H$ is not an inner product, an inner-product index cannot compute it. SparseCL therefore works in two stages (\cref{fig:sparsecl} in Appendix~\ref{app:extra}). First, an index retrieves the top 1,000 passages by cosine similarity. Then $F$ reranks these candidates. Empirically, this is almost equivalent to scoring the whole corpus. Scoring from stored embeddings is still much cheaper than using a cross-encoder, which SparseCL measured to be at least 200 times slower. The practical difference between the two methods lies here. \mc's score is itself an inner product, so it needs no second stage. In addition, each passage stores $768 + n(k+2)$ numbers instead of two 768-dimensional embeddings.

\section{First Round: SparseCL's Benchmarks}
\label{sec:round1}

\paragraph{Benchmarks and protocol.} We use SparseCL's benchmarks, data splits and evaluation protocol. The first benchmark is ArguAna \citep{wachsmuth2018retrieval}, a counter-argument retrieval task. The other two use passages from MSMARCO \citep{nguyen2016msmarco} and HotpotQA \citep{yang2018hotpotqa}, which GPT-4 rewrote into paraphrases and contradictions. In these two benchmarks, GPT-4 wrote several paraphrases and several contradictions of each original passage. A query is one of the paraphrases. Its targets are the contradictions of the same passage, and the other paraphrases of that passage are its confounders: they say the same thing as the query, so they are very similar to it, but they do not contradict it. The top 1,000 passages by cosine similarity are reranked, $\alpha$ is chosen by SparseCL's interval search on the development split, and we report NDCG@10 on the test split. For SparseCL, we run its released bge-base models with its own test script. This reproduces its published numbers (0.704, 0.909 and 0.966, against 0.704, 0.909 and 0.967). \mc\ uses the same zero-shot cosine encoder (bge-base-en-v1.5; \citealt{xiao2024cpack}), and we train three seeds per configuration.

\begin{table}[t]
\caption{First round: SparseCL's benchmarks (NDCG@10, test). SparseCL: its released bge models with its own test script. \mc: mean $\pm$ sd over three seeds. Numbers: stored per passage.}
\label{tab:round1}
\centering
\small
\setlength{\tabcolsep}{3pt}
\resizebox{\columnwidth}{!}{%
\begin{tabular}{lcccr}
\toprule
System & ArguAna & MSMARCO & HotpotQA & Numbers \\
\midrule
Zero-shot cosine & 0.657 & 0.600 & 0.597 & 768 \\
SparseCL & \textbf{0.704} & 0.909 & \textbf{0.966} & 1,536 \\
\mc, $16\times8$ & $0.686\pm\sd{0.002}$ & $\mathbf{0.927}\pm\sd{0.001}$ & $0.957\pm\sd{0.001}$ & 928 \\
\mc, $2\times1$ & $0.680\pm\sd{0.002}$ & $0.894\pm\sd{0.000}$ & $0.941\pm\sd{0.003}$ & 774 \\
\bottomrule
\end{tabular}}
\end{table}

\begin{table}[t]
\caption{The query-free control on SparseCL's benchmarks (NDCG@10, test). Query-free: every query's learned representation (sparse embedding, clash vector) replaced by the average one. Gain kept: the share of the gain over zero-shot cosine (\cref{tab:round1}) that survives without the query.}
\label{tab:queryfree}
\centering
\small
\setlength{\tabcolsep}{4pt}
\resizebox{\columnwidth}{!}{%
\begin{tabular}{lccc}
\toprule
System & ArguAna & MSMARCO & HotpotQA \\
\midrule
SparseCL & 0.704 & 0.909 & 0.966 \\
\quad query-free & 0.695 & 0.686 & 0.827 \\
\quad gain kept (\%) & 79 & 28 & 63 \\
\midrule
\mc, $16\times8$ & $0.686\pm\sd{0.002}$ & $0.927\pm\sd{0.001}$ & $0.957\pm\sd{0.001}$ \\
\quad query-free & $0.682\pm\sd{0.003}$ & $0.859\pm\sd{0.004}$ & $0.940\pm\sd{0.001}$ \\
\quad gain kept (\%) & $85\pm\sd{8}$ & $79\pm\sd{1}$ & $95\pm\sd{0}$ \\
\bottomrule
\end{tabular}}
\end{table}

\paragraph{Results.} \mc\ performs on par with SparseCL (\cref{tab:round1}). It is better on MSMARCO, slightly below on ArguAna and HotpotQA, and equal on average (0.857 against 0.860), with a 40\% smaller index. Exact search over the whole corpus gives the same results (Appendix~\ref{app:details}).

\paragraph{An unexpected finding.} To understand what \mc\ had learned, we read the passages with the most extreme values of individual slots (Appendix~\ref{app:slots}). One slot behaves as a polarity scale. It orders passages by how much negation or denial they contain. Another slot separates GPT-4's generated contradictions, with their placeholder names and deflating claims, from genuine descriptions. Both are properties of a single passage, not of the pair. In fact, \mc's stored vector for a passage, without any query, can tell the benchmarks' contradictions from their paraphrases. For the best single slot, the AUC is 0.86 on ArguAna, 0.91 on MSMARCO and 0.99 on HotpotQA. This suggested that the benchmarks might be largely solvable without comparing the query with the passage.

\paragraph{Query-free control.} To measure this, we replace every query's learned representation with the average over all test queries, for both methods. For \mc, this is the average clash vector $k(q)$. For SparseCL, it is the average sparse embedding $s(q)$. Cosine still compares the query with the passage, but the learned term now depends on the passage alone. We tune $\alpha$ on the development split under the same control. We then report the share of the gain over cosine that survives.

\paragraph{Both methods keep much of their gain without the query.} \mc\ keeps 79--95\% of its gain over cosine on all three benchmarks (\cref{tab:queryfree}). SparseCL keeps 79\% on ArguAna and 63\% on HotpotQA. SparseCL thus depends on the passage alone less than \mc\ does, most clearly on MSMARCO (28\% against 79\%). The clash is nonetheless a pairwise measure. Changing a single fact in a passage raises it 7 times (ArguAna) to 80--110 times (MSMARCO and HotpotQA) more than deleting an article does. Recognising what a contradicting passage looks like is simply an effective strategy on these benchmarks, for any method trained on them. SparseCL's authors anticipated one shortcut of this kind and removed it. GPT-4's passages are easy to tell apart from human-written ones, so they left the original human-written passages out of the corpora, and every passage in them is generated \citep{xu2025contradiction}. Within the generated text, however, a second shortcut remains. Contradictions are written to disagree with a passage, and paraphrases are written to agree with it, and the two can be told apart by their style alone. These benchmarks therefore cannot show which method compares a query with a passage better. We need a benchmark on which this strategy cannot work.

\section{A Benchmark without the Shortcut}
\label{sec:benchmark}

\begin{figure*}[t]
\centering
\includegraphics[width=0.85\textwidth]{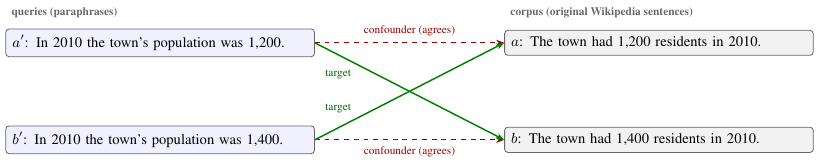}
\caption{The symmetric benchmark (illustrative sentences). For a pair of Wikipedia sentences $a$ and $b$ that contradict each other, both enter the corpus and their paraphrases become the queries. Each corpus sentence is the target of one query and the confounder of the other, so any preference that looks at passages alone helps one query exactly where it hurts the other.}
\label{fig:symmetric}
\end{figure*}

\paragraph{Construction.} We compile a new contradiction benchmark. VitaminC \citep{schuster2021vitaminc} collects real Wikipedia revisions in which a factual edit changes the verdict on a claim. The sentence before the edit supports the claim, and the sentence after it refutes it, or the other way round. We take each such pair $(a,b)$ of near-identical, human-written sentences that carry opposite verdicts for the same claim. We use only real revisions, and each sentence appears in at most one pair. Both sentences enter the corpus. We paraphrase them by back-translation \citep{mallinson2017paraphrasing} through German, using OPUS-MT models \citep{tiedemann2020opus}. The paraphrases $a'$ and $b'$ become the queries. Query $a'$ must retrieve $b$, with $a$ as its confounder, and query $b'$ must retrieve $a$, with $b$ as its confounder (\cref{fig:symmetric}). We keep a paraphrase only if it preserves every number and every capitalised word of its sentence, because these are the facts that VitaminC's edits change. We keep a pair only if both of its sentences get a paraphrase. The training triples have the form (query, contradiction, confounder), exactly as in SparseCL's data, and the files use SparseCL's format.

\paragraph{The property.} Every corpus passage is the target of exactly one query and the confounder of exactly one other query. Consider a score $f(d)$ that ignores the query. To answer $a'$, it must prefer $b$ over $a$. To answer $b'$, it must prefer $a$ over $b$. It can therefore answer at most one query of each pair. Cosine remains a strong confounder, because $a'$ is a paraphrase of $a$. Zero-shot cosine ranks a query's target above its confounder for only 1.2\% of test queries.

\paragraph{Data.} The benchmark has 17,978 training pairs (35,956 queries), 1,000 development pairs and 1,966 test pairs (3,932 test queries). As in SparseCL's benchmarks, the corpus is the union of all splits (41,888 passages). Automatic checks confirm the file format for both pipelines, the roles of every passage, and the preservation of numbers and names. They also confirm that simple features of a passage alone, such as negation words, length or digits, rank targets above confounders for at most half of the queries. Most edits change a number. Of the test queries, 61\% involve a changed number, 21\% a changed name, 17\% other wording (for instance, ``positive'' reviews becoming ``negative'' ones), and 1\% an added or removed negation. We classify these edits automatically from the changed words.

\section{Second Round: The Symmetric Benchmark}
\label{sec:round2}

\begin{table*}[t]
\caption{Second round: the symmetric benchmark (test split; mean $\pm$ sd over three seeds). Pairwise accuracy: the share of queries whose target scores above its confounder, the sentence the query paraphrases. Query-free: every query's learned representation replaced by the average one. Test-only corpus: the test split's passages only.}
\label{tab:round2}
\centering
\small
\begin{tabular}{lcccccc}
\toprule
& \multicolumn{2}{c}{NDCG@10 (full corpus)} & \multicolumn{2}{c}{Pairwise accuracy (\%)} & \multicolumn{2}{c}{NDCG@10 (test-only corpus)} \\
\cmidrule(lr){2-3}\cmidrule(lr){4-5}\cmidrule(lr){6-7}
System & normal & query-free & normal & query-free & normal & query-free \\
\midrule
Zero-shot cosine & 0.535 & -- & 1.2 & -- & 0.566 & -- \\
SparseCL & $0.569\pm\sd{0.002}$ & $0.535\pm\sd{0.001}$ & $12.1\pm\sd{1.3}$ & $1.4\pm\sd{0.2}$ & $0.604\pm\sd{0.003}$ & $0.559\pm\sd{0.008}$ \\
\mc, $16\times8$ & $\mathbf{0.621}\pm\sd{0.002}$ & $0.559\pm\sd{0.001}$ & $\mathbf{39.9}\pm\sd{1.2}$ & $12.5\pm\sd{1.4}$ & $\mathbf{0.668}\pm\sd{0.001}$ & $0.595\pm\sd{0.002}$ \\
\mc, $2\times1$ & $0.588\pm\sd{0.024}$ & $0.545\pm\sd{0.008}$ & $24.4\pm\sd{8.1}$ & $8.6\pm\sd{4.8}$ & $0.629\pm\sd{0.025}$ & $0.581\pm\sd{0.012}$ \\
\bottomrule
\end{tabular}
\end{table*}

\paragraph{Setup.} Both methods use bge-base. We train SparseCL's sparse model on the new training split with SparseCL's own code and settings: three epochs, batches of 64, a learning rate of $2\cdot10^{-5}$, a temperature of 0.02 and its sparsity loss. We evaluate it with its own test script, with $\alpha$ chosen by its search on the development split. \mc\ is trained and selected in the same way, with its own default settings. We train it in the two configurations of the first round and, for the capacity analysis below, in four more. Each configuration is trained with three seeds. Besides NDCG@10, we report two measures. The first is pairwise accuracy, the share of test queries whose target scores above its confounder. This is the benchmark's core decision. The second is NDCG@10 against a test-only corpus, which excludes the training passages that the trained models have seen.

\paragraph{The benchmark removes the shortcut for SparseCL.} SparseCL's query-free score falls back to the level of cosine (0.535; \cref{tab:round2}). On this benchmark, none of its gain comes from the passage alone.

\paragraph{\mc\ compares the pair far better.} With 16 slots, \mc\ improves NDCG@10 over cosine by 0.086, while SparseCL improves it by 0.034 (0.621 against 0.569; $p<0.001$, Welch's $t$-test over seeds). On the core decision, \mc\ ranks the contradiction above its confounder for 39.9\% of queries. This is more than three times SparseCL's 12.1\% ($p<0.001$), and far above the 1.2\% of cosine.

\paragraph{The gain survives the query-free subtraction.} Unlike SparseCL's, \mc's query-free score does not fall all the way back to cosine. It reaches 12.5\% pairwise accuracy. This is the expected effect of a preference over passages on near-ties. When cosine almost ties a target with its confounder, a preference between the two passages can tip one query of the pair the right way. By construction, it can never tip both. If we subtract each method's own query-free result, \mc\ still gains more than twice as much as SparseCL (27.4 against 10.7 points of pairwise accuracy).

\paragraph{It is not a memory of seen passages.} On a corpus of the test passages only, which the trained models have never seen, \mc's margin persists (0.668 against 0.604 for SparseCL and 0.566 for cosine; $p<0.001$), and so does its query-free gain (0.595).

\paragraph{Why SparseCL gains so little here.} SparseCL improves on cosine far less on this benchmark than on its own benchmarks. Before interpreting this, we checked that our way of training it is not to blame. We trained SparseCL with exactly the pipeline and settings we use here, but on its own HotpotQA data. It then reproduces its released model, with an NDCG@10 of 0.965 against 0.966 (\cref{tab:sparsecl-check} in Appendix~\ref{app:extra}). Its sparsity score also separates contradictions from confounders in the same way. On VitaminC, however, the same training yields a sparsity score that does not tell a contradiction from its confounder at all. The median Hoyer sparsity of the difference is 0.209 for the contradiction and 0.210 for the confounder. The score prefers the contradiction for only 44\% of queries, slightly less often than chance.

We believe that the reason lies in what the score measures. In our benchmark, a query is a paraphrase of one of the two sentences, so it differs from both sentences in many words. The single changed fact barely alters how concentrated this difference is. On SparseCL's own benchmarks, the difference between a query and its generated contradiction is concentrated in a few dimensions, while the difference to a paraphrase is spread out. The sparsity score was trained to exploit exactly this. \mc's values, in contrast, encode which numbers a sentence mentions, largely independently of how it is worded (\cref{sec:analysis}). A changed number therefore changes them, even under paraphrase.

\paragraph{What capacity \mc\ needs.} Appendix~\ref{app:capacity} varies the number of slots and the size of each value. What matters is the size of the values. With eight numbers per value, even two slots are enough. This configuration stores only 20 numbers per passage beyond the cosine embedding, about half of SparseCL's index. It ranks the contradiction first for 39.6\% of queries ($p<0.001$ against SparseCL).

\paragraph{The task remains open.} Even the best configurations still rank the confounder first for more than 60\% of queries.

\section{What \mc's Values Encode}
\label{sec:analysis}

The clash decomposes exactly into one term per slot. We can therefore analyse it slot by slot. We analyse the three trained \mc\ models with 16 slots of 8 numbers, on the test set of the symmetric benchmark.

\paragraph{The slots do not divide the work.} Every slot commits on every passage. A contradiction's clash is spread almost evenly across the slots. Thirteen of the 16 slots are needed to account for 80\% of it, and the largest single slot carries only 7\%. The slot that carries the most does not depend on the kind of edit (a number, a name, other wording or a negation). Their normalised mutual information is at the level of shuffled labels (0.003--0.004 against 0.004). The slot with the largest clash attends to the edited words no more than chance would predict (11\% of its attention, against 11\%). Switching off any single slot costs about the same, about 2 points of pairwise accuracy. Grouping each slot's values over the 41,888 sentences of the corpus confirms this. Within a model, the 16 slots organise the sentences in only one or two distinct ways. The slots thus behave as copies of one another, not as separate units that could be interpreted one by one.

\paragraph{The values fingerprint numbers.} What these copies encode is readable, however. When we cluster the values, the clusters group sentences by the numbers they mention. Numbers written as words fall together with their digits, such as ``seven'' with ``7'', or ``three'' and ``third'' with ``3''. Numbers ending in the same digit tend to group together, such as 4, 14 and 24, or 5, 25 and 35. Review verdicts such as ``negative'' and ``mixed'' appear among the distinctive words of some clusters. The same pattern appears in all three models.

\paragraph{A controlled probe.} To test this reading, we wrote five sentence templates, such as ``The town had \underline{\hspace{1.2em}} residents in 2010.'' and ``The team won \underline{\hspace{1.2em}} games that season.''. We filled each template with the numbers 1 to 100, written both as digits and as words, giving 1,000 sentences in total. We then compared their values using the model's own clash (\cref{tab:probe} in Appendix~\ref{app:extra}). The value of a sentence depends almost only on its number. The same number in two different templates clashes about 5\% as much as two different numbers do. A number written as a word lands closest to its own digits about a quarter of the time, which is 25 times more often than chance. The two forms of a number clash about half as much as different numbers do. Numbers that share their last digit clash about a quarter less than other pairs. The size of a number, however, plays no role at all. The clash between two numbers is unrelated to how far apart they are, so 6 against 7 clashes as much as 6 against 90.

\paragraph{What this explains.} The values act as fingerprints of the numbers in a sentence. They capture the form of a number rather than its size, and they are largely independent of the wording around it. This is exactly what the symmetric benchmark rewards. Most of its edits change a number, and a fingerprint that survives paraphrase registers such a change, whatever the rest of the query looks like. It also marks a limit of what the model knows. Since size plays no role, ``about 1,200'' and ``1,250'' look as contradictory to the model as ``1,200'' and ``9,000''.

\section{Related Work}
\label{sec:related}

\paragraph{Contradiction and stance retrieval.} SparseCL \citep{xu2025contradiction} introduced contradiction retrieval as a task, together with its benchmarks and the Hoyer-sparsity score. We build on its protocol, data and training objective. We contribute an inner-product score, a control and a new benchmark. Embedding models for stance-aware argument retrieval favour topical overlap over stance \citep{sparacino2026embedding}. Retrievers also rarely separate supporting from opposing documents for the same claim \citep{zhao2024beyond}. In fact verification, retrieving supporting and refuting evidence separately improves recall \citep{dong2021robust}.

\paragraph{Shortcuts and controls.} Hypothesis-only baselines revealed that NLI datasets can be partly solved without looking at the premise \citep{gururangan2018annotation,poliak2018hypothesis}. Similarly, FEVER's claims reveal their label without any evidence \citep{thorne2018fever,schuster2019towards}. Our query-free control is the counterpart of these baselines for retrieval. It asks how much of a contradiction score's gain survives without the query. Contrast sets \citep{gardner2020contrast} and counterfactually augmented data \citep{kaushik2020learning} evaluate models on inputs that differ minimally but have different labels. Our benchmark applies the same idea to retrieval, by making every passage both a target and a confounder. VitaminC \citep{schuster2021vitaminc} built contrastive evidence from Wikipedia revisions for fact verification, and we turn this evidence into retrieval pairs.

\paragraph{Indexable similarity.} Bi-encoders \citep{reimers2019sentencebert} and dense retrievers \citep{karpukhin2020dense} made inner-product search over learned embeddings standard. Libraries such as FAISS \citep{johnson2019billion} serve this search at scale. Late-interaction models \citep{khattab2020colbert} store one vector per token instead. \mc\ keeps a single stored vector per passage, but it makes the inner product compute a distance between learned values. This escapes the transitivity of similarity.

\section{Conclusion}
\label{sec:conclusion}

\mc\ turns contradiction retrieval into a single inner product. Any inner-product index can compute it exactly, without reranking and with a smaller index than SparseCL. On the standard benchmarks, the two methods perform on par, but much of what trained methods gain there is available from the passage alone. We therefore built a benchmark from real Wikipedia revisions, on which every passage is both a contradiction and a confounder. On it, \mc\ ranks contradictions above their confounders three times as often as SparseCL, with only two slots of eight numbers. SparseCL's sparsity score does not separate these real single-fact edits at all. \mc's values fingerprint the numbers a sentence mentions, by their form rather than their size. We recommend that evaluations of contradiction retrieval report a query-free control, on benchmarks that the style of passages cannot answer.

\section*{Limitations}
\paragraph{Noise in the benchmark.} We assume that a before/after pair is a contradiction because the two sentences carry opposite verdicts for the same claim. Some pairs may differ in ways that do not amount to a contradiction. Back-translation may also change the meaning in ways that our filter does not check, since the filter only checks numbers and names.

\paragraph{Training settings.} The results reported for both methods on the new benchmark use their default settings, with only $\alpha$ tuned on the development split. Our training of SparseCL reproduces its released model on HotpotQA. Its weak result on the new benchmark is therefore not a failure of our training under these settings, but other settings might still help it. SparseCL also offers a fine-tuned cosine encoder, which we did not use on the new benchmark.

\paragraph{Capacity.} \mc\ needs values of several numbers. With a single number per slot, it is only slightly better than SparseCL. We compared the configurations with eight numbers per value on the symmetric benchmark only. The first round used 16 slots of 8 numbers and 2 slots of 1 number.

\paragraph{Cross-encoders.} Like SparseCL, we compare methods that score a corpus from stored embeddings. A cross-encoder reads the query and the passage together. It can judge a given pair more accurately, but it cannot score a whole corpus from stored vectors, so it is outside this comparison.

\paragraph{Query-free control.} The control replaces a learned query representation with an average. Other ways of removing the query could give different shares. On the symmetric benchmark, a preference over passages can still help one query of a pair on near-ties. This is why we subtract each method's own query-free result.

\paragraph{What the model learns.} The slots of the trained models do not specialise. They act as interchangeable copies, so individual slots should not be read as interpretable units. The values encode numbers by their form, not their size. Approximately equal numbers therefore look as contradictory as very different ones. The analysis covers the $16\times8$ models on the symmetric benchmark. The readings of individual slots on SparseCL's benchmarks (Appendix~\ref{app:slots}) come from pilot models.

\paragraph{Form of the score.} Other asymmetric inner products of the same size may perform similarly. We do not claim that \mc's particular form is required for its results.

\section*{Ethics Statement}
Contradiction retrieval can surface passages that conflict with a claim, for example in fact checking. A retrieved contradiction should be treated as a candidate for human review, not as a verdict. Our benchmark is built from VitaminC's public Wikipedia revisions and from automatic paraphrases. It contains no personal data beyond what Wikipedia articles contain. The code and the benchmark are available at \url{https://github.com/semantic-systems/moldcast}.

\bibliography{refs}

@article{dong2021robust,
  title   = {Robust Information Retrieval for False Claims with Distracting Entities in Fact Extraction and Verification},
  author  = {Dong, Mingwen and Christodoulopoulos, Christos and Shih, Sheng-Min and Ma, Xiaofei},
  journal = {arXiv preprint arXiv:2112.07618},
  year    = {2021}
}

@article{erickson1981words,
  title   = {From Words to Meaning: A Semantic Illusion},
  author  = {Erickson, Thomas D. and Mattson, Mark E.},
  journal = {Journal of Verbal Learning and Verbal Behavior},
  volume  = {20},
  number  = {5},
  pages   = {540--551},
  year    = {1981}
}

@inproceedings{gardner2020contrast,
  title     = {Evaluating Models' Local Decision Boundaries via Contrast Sets},
  author    = {Gardner, Matt and Artzi, Yoav and Basmov, Victoria and Berant, Jonathan and Bogin, Ben and Chen, Sihao and others},
  booktitle = {Findings of the Association for Computational Linguistics: EMNLP 2020},
  pages     = {1307--1323},
  year      = {2020}
}

@inproceedings{gururangan2018annotation,
  title     = {Annotation Artifacts in Natural Language Inference Data},
  author    = {Gururangan, Suchin and Swayamdipta, Swabha and Levy, Omer and Schwartz, Roy and Bowman, Samuel R. and Smith, Noah A.},
  booktitle = {Proceedings of the 2018 Conference of the North American Chapter of the Association for Computational Linguistics},
  year      = {2018}
}

@article{hagoort2004integration,
  title   = {Integration of Word Meaning and World Knowledge in Language Comprehension},
  author  = {Hagoort, Peter and Hald, Lea and Bastiaansen, Marcel and Petersson, Karl Magnus},
  journal = {Science},
  volume  = {304},
  number  = {5669},
  pages   = {438--441},
  year    = {2004}
}

@article{hoyer2004nonnegative,
  title   = {Non-negative Matrix Factorization with Sparseness Constraints},
  author  = {Hoyer, Patrik O.},
  journal = {Journal of Machine Learning Research},
  volume  = {5},
  year    = {2004}
}

@article{johnson2019billion,
  title   = {Billion-Scale Similarity Search with {GPUs}},
  author  = {Johnson, Jeff and Douze, Matthijs and J{\'e}gou, Herv{\'e}},
  journal = {IEEE Transactions on Big Data},
  year    = {2019}
}

@inproceedings{karpukhin2020dense,
  title     = {Dense Passage Retrieval for Open-Domain Question Answering},
  author    = {Karpukhin, Vladimir and O{\u{g}}uz, Barlas and Min, Sewon and Lewis, Patrick and Wu, Ledell and Edunov, Sergey and Chen, Danqi and Yih, Wen-tau},
  booktitle = {Proceedings of the 2020 Conference on Empirical Methods in Natural Language Processing},
  year      = {2020}
}

@inproceedings{kaushik2020learning,
  title     = {Learning the Difference that Makes a Difference with Counterfactually-Augmented Data},
  author    = {Kaushik, Divyansh and Hovy, Eduard and Lipton, Zachary C.},
  booktitle = {International Conference on Learning Representations},
  year      = {2020}
}

@inproceedings{khattab2020colbert,
  title     = {{ColBERT}: Efficient and Effective Passage Search via Contextualized Late Interaction over {BERT}},
  author    = {Khattab, Omar and Zaharia, Matei},
  booktitle = {Proceedings of the 43rd International ACM SIGIR Conference on Research and Development in Information Retrieval},
  year      = {2020}
}

@article{kutas1980reading,
  title   = {Reading Senseless Sentences: Brain Potentials Reflect Semantic Incongruity},
  author  = {Kutas, Marta and Hillyard, Steven A.},
  journal = {Science},
  volume  = {207},
  number  = {4427},
  pages   = {203--205},
  year    = {1980}
}

@inproceedings{mallinson2017paraphrasing,
  title     = {Paraphrasing Revisited with Neural Machine Translation},
  author    = {Mallinson, Jonathan and Sennrich, Rico and Lapata, Mirella},
  booktitle = {Proceedings of the 15th Conference of the European Chapter of the Association for Computational Linguistics: Volume 1, Long Papers},
  pages     = {881--893},
  year      = {2017}
}

@article{myers1998accessing,
  title   = {Accessing the Discourse Representation during Reading},
  author  = {Myers, Jerome L. and O'Brien, Edward J.},
  journal = {Discourse Processes},
  volume  = {26},
  number  = {2--3},
  pages   = {131--157},
  year    = {1998}
}

@inproceedings{nguyen2016msmarco,
  title     = {{MS MARCO}: A Human Generated Machine Reading Comprehension Dataset},
  author    = {Nguyen, Tri and Rosenberg, Mir and Song, Xia and Gao, Jianfeng and Tiwary, Saurabh and Majumder, Rangan and Deng, Li},
  booktitle = {Proceedings of the Workshop on Cognitive Computation: Integrating Neural and Symbolic Approaches (CoCo@NIPS)},
  year      = {2016}
}

@article{obrien2016coherence,
  title   = {Coherence Threshold and the Continuity of Processing: The {RI-Val} Model of Comprehension},
  author  = {O'Brien, Edward J. and Cook, Anne E.},
  journal = {Discourse Processes},
  volume  = {53},
  number  = {5--6},
  pages   = {326--338},
  year    = {2016}
}

@inproceedings{poliak2018hypothesis,
  title     = {Hypothesis Only Baselines in Natural Language Inference},
  author    = {Poliak, Adam and Naradowsky, Jason and Haldar, Aparajita and Rudinger, Rachel and Van Durme, Benjamin},
  booktitle = {Proceedings of the Seventh Joint Conference on Lexical and Computational Semantics},
  year      = {2018}
}

@inproceedings{reimers2019sentencebert,
  title     = {Sentence-{BERT}: Sentence Embeddings using {Siamese} {BERT}-Networks},
  author    = {Reimers, Nils and Gurevych, Iryna},
  booktitle = {Proceedings of the 2019 Conference on Empirical Methods in Natural Language Processing and the 9th International Joint Conference on Natural Language Processing},
  year      = {2019}
}

@article{richter2009you,
  title   = {You Don't Have to Believe Everything You Read: Background Knowledge Permits Fast and Efficient Validation of Information},
  author  = {Richter, Tobias and Schroeder, Sascha and W{\"o}hrmann, Britta},
  journal = {Journal of Personality and Social Psychology},
  volume  = {96},
  number  = {3},
  pages   = {538--558},
  year    = {2009}
}

@inproceedings{schuster2021vitaminc,
  title     = {Get Your Vitamin {C}! Robust Fact Verification with Contrastive Evidence},
  author    = {Schuster, Tal and Fisch, Adam and Barzilay, Regina},
  booktitle = {Proceedings of the 2021 Conference of the North American Chapter of the Association for Computational Linguistics: Human Language Technologies},
  pages     = {624--643},
  year      = {2021}
}

@inproceedings{schuster2019towards,
  title     = {Towards Debiasing Fact Verification Models},
  author    = {Schuster, Tal and Shah, Darsh J. and Yeo, Yun Jie Serene and Filizzola, Daniel and Santus, Enrico and Barzilay, Regina},
  booktitle = {Proceedings of the 2019 Conference on Empirical Methods in Natural Language Processing and the 9th International Joint Conference on Natural Language Processing},
  year      = {2019}
}

@inproceedings{sparacino2026embedding,
  title     = {Embedding Models for Stance-Aware Argument Retrieval},
  author    = {Sparacino, Angelo and Dejl, Adam},
  booktitle = {Proceedings of the Workshop on Computational Models of Natural Argument (CMNA)},
  note      = {ArXiv:2608.28283},
  year      = {2026}
}

@inproceedings{thorne2018fever,
  title     = {{FEVER}: a Large-scale Dataset for Fact Extraction and {VER}ification},
  author    = {Thorne, James and Vlachos, Andreas and Christodoulopoulos, Christos and Mittal, Arpit},
  booktitle = {Proceedings of the 2018 Conference of the North American Chapter of the Association for Computational Linguistics},
  year      = {2018}
}

@inproceedings{tiedemann2020opus,
  title     = {{OPUS-MT} -- Building Open Translation Services for the World},
  author    = {Tiedemann, J{\"o}rg and Thottingal, Santhosh},
  booktitle = {Proceedings of the 22nd Annual Conference of the European Association for Machine Translation},
  pages     = {479--480},
  year      = {2020}
}

@inproceedings{wachsmuth2018retrieval,
  title     = {Retrieval of the Best Counterargument without Prior Topic Knowledge},
  author    = {Wachsmuth, Henning and Syed, Shahbaz and Stein, Benno},
  booktitle = {Proceedings of the 56th Annual Meeting of the Association for Computational Linguistics},
  year      = {2018}
}

@inproceedings{xiao2024cpack,
  title     = {{C-Pack}: Packed Resources for General {Chinese} Embeddings},
  author    = {Xiao, Shitao and Liu, Zheng and Zhang, Peitian and Muennighoff, Niklas and Lian, Defu and Nie, Jian-Yun},
  booktitle = {Proceedings of the 47th International ACM SIGIR Conference on Research and Development in Information Retrieval},
  year      = {2024}
}

@inproceedings{xu2025contradiction,
  title     = {Contradiction Retrieval via Contrastive Learning with Sparsity},
  author    = {Xu, Haike and Lin, Zongyu and Chang, Kai-Wei and Sun, Yizhou and Indyk, Piotr},
  booktitle = {Proceedings of the 42nd International Conference on Machine Learning},
  series    = {Proceedings of Machine Learning Research},
  volume    = {267},
  year      = {2025}
}

@inproceedings{yang2018hotpotqa,
  title     = {{HotpotQA}: A Dataset for Diverse, Explainable Multi-hop Question Answering},
  author    = {Yang, Zhilin and Qi, Peng and Zhang, Saizheng and Bengio, Yoshua and Cohen, William W. and Salakhutdinov, Ruslan and Manning, Christopher D.},
  booktitle = {Proceedings of the 2018 Conference on Empirical Methods in Natural Language Processing},
  year      = {2018}
}

@article{zhao2024beyond,
  title   = {Beyond Relevance: Evaluate and Improve Retrievers on Perspective Awareness},
  author  = {Zhao, Xinran and Chen, Tong and Chen, Sihao and Zhang, Hongming and Wu, Tongshuang},
  journal = {arXiv preprint arXiv:2405.02714},
  year    = {2024}
}
\bibliographystyle{icml2026}

\newpage
\appendix

\section{Training and Evaluation Details}
\label{app:details}

\paragraph{Clash models.} The slot encoder is initialised from bge-base-en-v1.5. It is followed by a slot head of width 256 with 2 mixing layers. Commitments are sigmoids, and values are linear. We train for three epochs, with batches of 64 triples and AdamW with a learning rate of $2\cdot10^{-5}$, a weight decay of 0.01 and 10\% warmup. A hinge loss keeps each passage's mean commitment above a floor, so that the slots cannot all fall silent. The zero-shot cosine embeddings come from the frozen bge-base model. We choose $\alpha$ on the development split with SparseCL's interval search over $[0,10]$, and we reuse it on the test split. Every evaluation also reports exact search over the whole corpus. On SparseCL's benchmarks, this gives the same results as the shortlist.

\paragraph{SparseCL.} We use SparseCL's code, in an environment with its pinned library versions. We record the exact commit with our release. On its own benchmarks, we run its released bge models. On the symmetric benchmark, we train its sparse model with the settings of its bge scripts. Its query-free control adds three lines to its test script, which replace every query's sparse embedding with the average one. The diagnostics of \cref{sec:round2} apply its scoring rule to the embeddings that its own script computes.

\paragraph{Benchmark construction.} We use VitaminC's train, validation and test splits as our training, development and test splits. We keep at most 20,000, 1,000 and 2,000 pairs from them. Paraphrases are generated with the OPUS-MT English--German and German--English models, using beam search with 4 beams. A paraphrase is kept if it differs from its sentence and preserves its numbers and capitalised words. More precisely, it must preserve the multiset of numbers and the set of capitalised words, except a capitalised word at the start of the sentence.

\section{Reading the Slots}
\label{app:slots}

We read the passages with the highest and lowest values of individual slots. The models are \mc\ models trained on SparseCL's benchmarks. They are pilot models with the same configurations as the final ones.

\paragraph{A polarity scale.} In a model with one number per slot, one slot orders passages by how much negation or denial they contain. At one end, we find passages such as ``It is not a Grade I listed building'' or ``He has never been named to an All-Star team and has not won any Gold Glove Awards''. At the other end, we find plain affirmative statements, such as ``She won the singles titles at the 1974 French Open''.

\paragraph{A direction for generated contradictions.} In the model with vector values, the main direction of one slot separates GPT-4's generated contradictions from genuine descriptions. At one end, we find passages with placeholder names (``by American folk artist John Doe, produced by Jane Smith''), deflating claims (``remained obscure, never entering the Top 100'') and implausible facts. At the other end, we find accurate descriptions, such as ``Jonathan Vincent Voight, born December 29, 1938''.

\section{Additional Figure and Tables}
\label{app:extra}

\begin{figure}[h]
\centering
\includegraphics[width=\columnwidth]{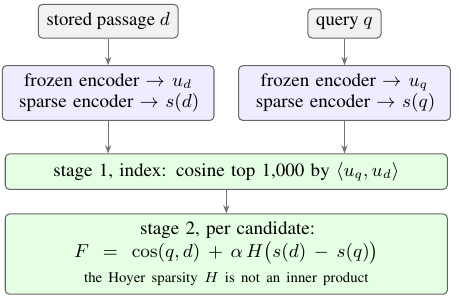}
\caption{SparseCL \citep{xu2025contradiction}, drawn like \mc\ in \cref{fig:pipeline}. A passage is encoded into a cosine and a sparse embedding (1,536 numbers). An index retrieves the cosine top 1,000; the Hoyer sparsity of the difference of sparse embeddings, which is not an inner product, reranks them.}
\label{fig:sparsecl}
\end{figure}

\Cref{fig:sparsecl} shows SparseCL's two-stage pipeline (\cref{sec:baseline}). \Cref{tab:sparsecl-check} checks our training of SparseCL (\cref{sec:round2}). \Cref{tab:probe} gives the per-seed results of the number probe (\cref{sec:analysis}). \Cref{tab:terms} compares each term of \mc's score alone and combined (\cref{sec:method}).

\begin{table}[h]
\caption{Checking our training of SparseCL. On its own HotpotQA benchmark, our training reproduces its released model (published NDCG@10: 0.967), and its sparsity score separates contradictions from confounders. On VitaminC, the same training yields no separation. Hoyer sparsity: median over test queries of the Hoyer sparsity of the difference between the sparse embeddings of the query and of its contradiction, or of its confounder (on HotpotQA, the query's most similar passage that is not a contradiction). VitaminC: means over three seeds.}
\label{tab:sparsecl-check}
\centering
\small
\setlength{\tabcolsep}{4pt}
\resizebox{\columnwidth}{!}{%
\begin{tabular}{llccc}
\toprule
& & & \multicolumn{2}{c}{Hoyer sparsity} \\
\cmidrule(lr){4-5}
Data & Model & NDCG@10 & contradiction & confounder \\
\midrule
HotpotQA & released & 0.966 & 0.631 & 0.281 \\
HotpotQA & our training & 0.965 & 0.632 & 0.283 \\
VitaminC & our training & 0.569 & 0.209 & 0.210 \\
\bottomrule
\end{tabular}}
\end{table}

\begin{table}[h]
\caption{How \mc\ ($16\times8$) encodes numbers: five templates, numbers 1--100 as digits and as words. Clash ratios compare the mean clash of the pairs described with that of pairs of different numbers; below 1 means closer.}
\label{tab:probe}
\centering
\small
\setlength{\tabcolsep}{3pt}
\resizebox{\columnwidth}{!}{%
\begin{tabular}{lccc}
\toprule
Measure & seed 0 & seed 1 & seed 2 \\
\midrule
Same number, other template (clash ratio) & 0.06 & 0.05 & 0.05 \\
Word finds its digit (top-1; chance 1\%) & 25\% & 26\% & 27\% \\
``six'' against ``6'' (clash ratio) & 0.50 & 0.50 & 0.49 \\
Same last digit, e.g.\ 6 and 16 (clash ratio) & 0.75 & 0.74 & 0.74 \\
Clash against distance in size (Spearman) & $-0.06$ & $-0.06$ & $-0.05$ \\
\bottomrule
\end{tabular}}
\end{table}

\begin{table}[h]
\caption{Each term of \mc's score alone and combined (NDCG@10, test; means over three seeds). Clash alone: passages ranked by the clash only, within the cosine shortlist of 1,000 passages or over the whole corpus. The clash is trained as a complement to cosine: on its own, it ranks unrelated passages first.}
\label{tab:terms}
\centering
\small
\setlength{\tabcolsep}{4pt}
\resizebox{\columnwidth}{!}{%
\begin{tabular}{llcccc}
\toprule
& & & \multicolumn{2}{c}{Clash alone} & \\
\cmidrule(lr){4-5}
Benchmark & Model & Cosine & shortlist & corpus & Cosine + clash \\
\midrule
ArguAna & $16\times8$ & 0.657 & 0.001 & 0.000 & 0.686 \\
        & $2\times1$  & 0.657 & 0.002 & 0.000 & 0.680 \\
MSMARCO & $16\times8$ & 0.600 & 0.027 & 0.000 & 0.927 \\
        & $2\times1$  & 0.600 & 0.013 & 0.000 & 0.894 \\
HotpotQA & $16\times8$ & 0.597 & 0.057 & 0.000 & 0.957 \\
         & $2\times1$  & 0.597 & 0.022 & 0.000 & 0.941 \\
VitaminC & $16\times8$ & 0.535 & 0.009 & 0.000 & 0.621 \\
         & $2\times1$  & 0.535 & 0.008 & 0.000 & 0.588 \\
\bottomrule
\end{tabular}}
\end{table}

\section{Capacity Analysis}
\label{app:capacity}

\mc's capacity has two dimensions: the number of slots and the size of each value. \Cref{tab:capacity} varies both. It shows that what matters is the size of the values. With a single number per value, \mc\ is weak whatever its number of slots. Its pairwise accuracy stays between 16\% and 24\%, only slightly above SparseCL's 12\%, and the configuration with two slots varies strongly from seed to seed. With eight numbers per value, every configuration wins clearly and almost equally. They reach an NDCG@10 of 0.621--0.624 whether they have 2, 16 or 64 slots, and these differences are not significant. On pairwise accuracy, 2 and 16 slots reach 39.6\% and 39.9\%, while 64 slots reach somewhat less, 35.6\% ($p=0.02$ against 16 slots). One number per slot thus appears to be too little to keep apart the many values a slot has to hold, while eight numbers are enough. The smallest of these configurations has two slots of eight numbers. It stores only 20 numbers per passage beyond the cosine embedding, 788 in total, which is about half of SparseCL's 1,536. It ranks the contradiction above its confounder for 39.6\% of queries, more than three times as often as SparseCL ($p<0.001$).

\begin{table}[h]
\caption{How much capacity \mc\ needs, on the symmetric benchmark (test; mean $\pm$ sd over three seeds). Configurations: slots $\times$ numbers per value. Numbers: stored per passage, $768 + \text{slots}\times(\text{numbers}+2)$. Pairwise accuracy as in \cref{tab:round2}.}
\label{tab:capacity}
\centering
\small
\setlength{\tabcolsep}{3pt}
\resizebox{\columnwidth}{!}{%
\begin{tabular}{lrccc}
\toprule
& & & \multicolumn{2}{c}{Pairwise accuracy (\%)} \\
\cmidrule(lr){4-5}
System & Numbers & NDCG@10 & normal & query-free \\
\midrule
\mc, $2\times1$  & 774 & $0.588\pm\sd{0.024}$ & $24.4\pm\sd{8.1}$ & $8.6\pm\sd{4.8}$ \\
\mc, $16\times1$ & 816 & $0.567\pm\sd{0.002}$ & $15.7\pm\sd{1.6}$ & $12.1\pm\sd{1.8}$ \\
\mc, $64\times1$ & 960 & $0.569\pm\sd{0.004}$ & $16.3\pm\sd{1.4}$ & $12.7\pm\sd{0.9}$ \\
\midrule
\mc, $2\times8$  & 788 & $\mathbf{0.624}\pm\sd{0.003}$ & $39.6\pm\sd{2.4}$ & $12.4\pm\sd{0.7}$ \\
\mc, $16\times8$ & 928 & $0.621\pm\sd{0.002}$ & $\mathbf{39.9}\pm\sd{1.2}$ & $12.5\pm\sd{1.4}$ \\
\mc, $64\times8$ & 1,408 & $0.621\pm\sd{0.003}$ & $35.6\pm\sd{1.6}$ & $13.5\pm\sd{0.3}$ \\
\midrule
SparseCL & 1,536 & $0.569\pm\sd{0.002}$ & $12.1\pm\sd{1.3}$ & $1.4\pm\sd{0.2}$ \\
\bottomrule
\end{tabular}}
\end{table}

\section{Contradiction Retrieval and Human Comprehension}
\label{app:human}

\paragraph{A disclaimer first.} This paper uses no brain, EEG or behavioural data. Nothing in it tests a claim about the brain. The parallel below concerns the computational problem itself, that is, what must be computed and under which constraints. It does not concern how the brain computes it. We mention the parallel because it motivates fast, large-scale contradiction retrieval as a problem worth studying.

\paragraph{Readers notice conflicts with what they know, and they do so quickly.} The N400 is a component of the event-related brain potential that peaks about 400\,ms after a word. It was first described for sentence endings that make no sense \citep{kutas1980reading}. Some sentences are well formed but conflict with world knowledge, such as ``Dutch trains are white'', when Dutch trains are in fact yellow. Such sentences elicit an N400 with the same timing as sentences that violate word meaning. This suggests that world knowledge is brought to bear as quickly as word meaning \citep{hagoort2004integration}.

\paragraph{The checking appears to be routine.} Behavioural experiments indicate that readers validate incoming information against their background knowledge, even when they are not asked to evaluate it. False statements such as ``Soft soap is edible'' are rejected efficiently, even under additional cognitive load \citep{richter2009you}.

\paragraph{Retrieval, then validation.} Memory-based models of comprehension describe how incoming information activates related content anywhere in long-term memory. This happens through a passive, resonance-like process, without a deliberate search \citep{myers1998accessing}. In the RI-Val model, activation, integration and validation are distinct passive processes. Validation is a pattern-matching check of the incoming information against what has been activated \citep{obrien2016coherence}. Read computationally, this is retrieval of relevant and possibly conflicting content from a very large store, followed by a check. This paper addresses the same task with one inner product per stored statement.

\paragraph{And it fails in telling ways.} Consider the question ``How many animals of each kind did Moses take on the Ark?''. Most people answer ``two'' without noticing that it was Noah, although they know it. The illusion largely disappears when the substituted name is unrelated, such as ``Nixon'' \citep{erickson1981words}. A conflict is thus easiest to miss when the conflicting statement is otherwise very similar to what is known. This is also why contradiction retrieval is hard for retrieval based on similarity. It is also why our benchmark pairs every contradiction with a near-identical confounder.

\paragraph{Where the parallel ends.} Human validation draws on far richer knowledge and reasoning than any model here. Our own analysis (\cref{sec:analysis}) shows one concrete difference. \mc\ encodes numbers by their form, not their size. It therefore treats ``1,250'' as contradicting ``about 1,200'' as strongly as ``9,000'' does, which a human reader would not. Nor do we suggest that the brain computes inner products, slots or clashes. What the comparison does suggest is a design target. Human comprehension appears to check every new statement against a large store, quickly and in a single pass. Computational models should aim for the same.

\end{document}